\pdfoutput=1

\documentclass[11pt]{article}
\usepackage{EMNLP2023}

\usepackage{times}
\usepackage{latexsym}

\usepackage[T1]{fontenc}

\usepackage[utf8]{inputenc}
\usepackage[T2A,LAE,T1]{fontenc}
\usepackage[arabic,USenglish]{babel}
\usepackage{CJKutf8}

\usepackage{microtype}

\usepackage{inconsolata}

\usepackage{booktabs}
\usepackage{graphicx}
\usepackage{multirow}
\usepackage{subcaption}

\usepackage{todonotes}
\usepackage{amsmath}
\usepackage[shortlabels]{enumitem}

\usepackage{tikz}
\usetikzlibrary{tikzmark}

\usepackage{xcolor}
\definecolor{correct_FP}{RGB}{146, 197, 222}
\definecolor{incorrect_FP}{RGB}{202, 0, 32}

\title{Arabic Dialect Identification under Scrutiny: Limitations of Single-label Classification}

\author{Amr Keleg \and Walid Magdy\\
Institute for Language, Cognition and Computation \\
School of Informatics, University of Edinburgh \\
\texttt{akeleg@sms.ed.ac.uk}, \texttt{wmagdy@inf.ed.ac.uk}}

\begin{document}
\maketitle
\begin{abstract}
Automatic Arabic Dialect Identification (ADI) of text has gained great popularity since it was introduced in the early 2010s. Multiple datasets were developed, and yearly shared tasks have been running since 2018. However, ADI systems are reported to fail in distinguishing between the micro-dialects of Arabic. We argue that the currently adopted framing of the ADI task as a single-label classification problem is one of the main reasons for that. We highlight the limitation of the incompleteness of the \textit{Dialect} labels and demonstrate how it impacts the evaluation of ADI systems. A manual error analysis for the predictions of an ADI, performed by 7 native speakers of different Arabic dialects, revealed that $\approx 66\%$ of the validated errors are not true errors. Consequently, we propose framing ADI as a multi-label classification task and give recommendations for designing new ADI datasets.
\end{abstract}

\section{Introduction}

ADI of text is an NLP task meant to determine the Arabic Dialect of the text from a predefined set of dialects. Arabic dialects can be grouped according to different levels (1) major regional level: Levant, Nile Basin, Gulf, Gulf of Aden, and Maghreb (2) country level: more than 20 Arab countries, and (3) city level: more than 100 micro-dialects \cite{cotterell-callison-burch-2014-multi, baimukan-etal-2022-hierarchical}.

Different datasets were built curating data from various resources with labels of different degrees of granularities: (1) regional-level \cite{zaidan-callison-burch-2011-arabic, alsarsour-etal-2018-dart}, (2) country-level \cite{abdelali-etal-2021-qadi, abdul-mageed-etal-2022-nadi, abdul-mageed-etal-2023-nadi}, or (3) city-level \cite{bouamor-etal-2019-madar, abdul-mageed-etal-2020-nadi, abdul-mageed-etal-2021-nadi}. Despite attracting lots of attention and effort for over a decade, ADI is still considered challenging, especially for the fine-grained distinction of micro-Arabic dialects on the country and city levels. This is generally demonstrated by the inability of ADI models to achieve high macro-F1 scores.

\begin{table}[t]
    \centering
    \small
    \resizebox{.5\textwidth}{!}{
        \begin{tabular}{p{3.5cm}r}
             \textbf{Dialects} & \textbf{Sentence} \\
             \midrule
             Iraq, Jordan, Lebanon, & \AR{وين المحطة؟} \\
             Libya, Oman, Palestine & \textbf{Where is the station?}\\
             Qatar, Saudi Arabia, Sudan, Syria, Tunisia, Yemen & \\
             \midrule
             Iraq, Morocco, Qatar & \AR{شنو رقم الرحلة؟}\\
             & \textbf{What is the flight/trip number?}\\
             \bottomrule
        \end{tabular}%
    }
    \caption{The MADAR corpus \cite{bouamor-etal-2018-madar} has English/French sentences manually translated into different Arabic dialects. The table shows two sentences having the same translation across multiple country-level dialects.}
    \label{tab:examples}
\end{table}

We believe that framing ADI as a single-label classification problem is a major limitation, especially for short sentences that might not have enough distinctive cues of a specific dialect as per Table \ref{tab:examples}. Therefore, assigning a \textbf{single dialect label} to each sentence either automatically (e.g.: using geotagging) or manually makes the labels incomplete, which in turn affects the fairness of the evaluation process. The single-label limitation for DI was also discussed for other languages such as French \cite{bernier-colborne-etal-2023-dialect}.

The need for improving the framing of ADI and consequently the ADI resources was previously noted by \citet{althobaiti2020automatic}, who concluded the \textit{Future Directions} section of her survey of Arabic Dialect Identification (ADI) with the following:
\begin{quote}
    ``There is also a need to criticize the available resources and analyze them in order to find the gaps in the available ADI resources."
\end{quote}

In this paper, we introduce the concept of \textit{Maximal Accuracy} for ADI datasets having single labels. We then provide recommendations for how to build new ADI datasets in a multi-label setup to alleviate the limitations of single-label datasets. We hope that our study will spark discussions among the Arabic NLP community about the modeling of the ADI task, which would optimally lead to the creation of new datasets of more complete labels, and help in improving the quality of the ADI models. The main contributions of the paper can be summarized as follows:

\begin{enumerate}[nosep]
    \item[1.] Criticizing the current modeling of the ADI task as a single-label classification task by empirically estimating the \textit{Maximal Accuracy} for multiple existing ADI datasets.
    \item[2.] Performing an error analysis for an ADI model by recruiting native speakers of seven different country-level Arabic dialects.
    \item[3.] Presenting a detailed proposal for how multi-label classification can be used for ADI.
\end{enumerate}

\begin{table*}[!h]
    \centering
    \small
    \resizebox{\textwidth}{!}{%
    \begin{tabular}{p{5.5cm}lp{8.5cm}}
         \textbf{Dataset} & \multicolumn{1}{c}{\textbf{Ct/Cn/Re}} & \multicolumn{1}{c}{\textbf{Description}} \\
         \midrule
         \textbf{(1) Manual Labeling} & & \\
         AOC \cite{zaidan-callison-burch-2011-arabic} & - / - / 5 * & - Online comments to news articles, manually labeled three times by crowd-sourced human annotators.  \\
         Facebook test set \cite{huang-2015-improved} & - / - / 3 & \multirow{2}{8.5cm}{- 2,382 public Facebook posts manually annotated into Egyptian, Levantine, Gulf Arabic, and MSA.} \\
         \begin{scriptsize}{Note: Data attached to the paper on ACL Anthology.}\end{scriptsize} &  & \\
         VarDial 2016 \cite{malmasi-etal-2016-discriminating} & - / - / 4 & \multirow{4}{8.5cm}{- Sentences sampled from transcripts of broadcast, debate and discussion programs from AlJazeera. The dialects of these recorded programs were manually labeled. MSA is included as a 5\textsuperscript{th} dialect class for the models. Audio features were used in the 2017 and 2018 editions to allow for building multimodal models.}\\
         \begin{scriptsize}Note: The link provided is not working.\end{scriptsize} & & \\
         VarDial 2017 \cite{zampieri-etal-2017-findings} & - / - / 4  &  \\
         VarDial 2018 \cite{zampieri-etal-2018-language} & - / - / 4  &  \\
         \begin{scriptsize}Note: VarDial 2018 used the same data as VarDial 2017.\end{scriptsize} & & \\
         ArSarcasm-v2 \cite{abu-farha-etal-2021-overview} & - / - / 4 * & - 15,548 tweets sampled from previous sentiment analysis datasets, annotated for their dialect (including MSA).\\
         \midrule
         \textbf{(2) Translation} & & \\
         Tatoeba \cite{ho2016tatoeba} & - / 8 / 4 & - An ever-growing crowdsourced corpus of multilingual translations, that include MSA and 8 different Arabic dialects.\\
         MPCA \cite{bouamor-etal-2014-multidialectal} & - / 5 / 3 & - 2,000 Egyptian Arabic sentences from a pre-existing corpus, manually translated into 4 other country-level dialects in addition to MSA. \\
         PADIC \cite{meftouh-etal-2015-machine} & 5 / 4 / 2 & - 6,400 sentences sampled from the transcripts of recorded conversations and movie/TV shows in Algerian Arabic and manually translated into 4 other dialects and MSA. \\
         DIAL2MSA \cite{MUBARAK18.13} & - / - / 4 & - Dialectal tweets manually translated into MSA. \\
         MADAR6 \cite{bouamor-etal-2019-madar} & 5 / 5 / 4 & - 10,000 sentences manually translated into 5 city-level Arabic dialects in addition to MSA. \\
         MADAR26 \cite{bouamor-etal-2019-madar} & 25 / 15 / 5 & - 2,000 sentences manually translated into 25 city-level Arabic dialects in addition to MSA. \\
         \midrule
         \textbf{(3) Distinctive Lexical Cues} & & \\
         DART \cite{alsarsour-etal-2018-dart}& - / - / 5 * & - Tweets streamed using a seed list of distinctive dialectal terms, which are used to initially assign a dialect to each tweet, before having them manually verified by crowdsourced annotators. \\
         Twt15DA \cite{althobaiti_2022}& - / 15 / 5 & \multirow{2}{8.5cm}{- Tweets curated by iteratively augmenting lists of distinctive dialectal cues, starting with a seed list for each dialect.} \\
         \begin{scriptsize}{Note: Data shared as (tweet IDs, labels) only.}\end{scriptsize} &  & \\
         \midrule
         \textbf{(4) Geo-tagging} & & \\
         \cite{mubarak-darwish-2014-using} & - / ? / ? & \multirow{2}{8.5cm}{- Arabic tweets streamed from Twitter, then automatically annotated using the reported user locations of the tweets' authors.}\\
         \begin{scriptsize}{Note: Not publicly available.}\end{scriptsize} &  & \\
         YouDACC \cite{salama-etal-2014-youdacc} & - / 8 / 5 * & \multirow{2}{8.5cm}{- Comments to youtube videos labeled using the videos' countries of origin, and the authors' locations.} \\
         \begin{scriptsize}{Note: Not publicly available.}\end{scriptsize} &  & \\
         OMCCA \cite{al2016opinion} & 5 / 2 / 2 & - 27,912 reviews scrapped from \url{Jeeran.com}, and automatically labeled using the location of the reviewer.\\
         MASC \cite{doi:10.1177/0165551516683908} & - / 6 / 4 & - 9,141 reviews curated from online reviewing sites, Google Play, Twitter, and Facebook. The country of the reviewer is used as a proxy for the dialect of the review. \\
         Shami \cite{abu-kwaik-etal-2018-shami} & - / 4 / 1 & - Sentences in one of the 4 Levantine dialects: (1) manually collected from discussions about public figures on online fora; (2) automatically collected from the Twitter timelines of public figures. \\
         ARAP-Tweet \cite{zaghouani-charfi-2018-arap} & - / 16 / 5 * & \multirow{2}{8.5cm}{- A corpus of tweets from 1100 users, annotated at the user level for the dialect, age, and gender.} \\
         \begin{scriptsize}{Note: No download link on their site.}\end{scriptsize} &  & \\
         ARAP-Tweet 2.0 \cite{charfi-etal-2019-fine} & - / 17 / 5 * & \multirow{2}{8.5cm}{- A corpus of tweets from about 3000 users, annotated at the user level for the dialect, age, and gender.} \\
         \begin{scriptsize}{Note: No download link on their site.}\end{scriptsize} &  & \\
         Habibi \cite{el-haj-2020-habibi} & - / 18 / 6 *$\dag$ & - Songs' lyrics labeled by the country of origin of their singers. \\
         QADI \cite{abdelali-etal-2021-qadi} & - / 18 / 5 & \multirow{3}{8.5cm}{- Tweets automatically labeled based on the locations of the authors in the user description field. The labels of the testing set of each country were validated by a native speaker of each country's dialect.}\\
         \begin{scriptsize}{Note: Training data shared as (tweet IDs, labels) only.}\end{scriptsize} &  & \\
         & & \\
         NADI2020 \cite{abdul-mageed-etal-2020-nadi} & 100 / 21 / 5 & \multirow{3}{8.5cm}{- Tweets of users staying in the same province for 10 months, automatically labeled by geotagging the tweets of the selected users.}\\
         NADI2021 \cite{abdul-mageed-etal-2021-nadi} & 100 / 21 / 5 & \\
         NADI2022 \cite{abdul-mageed-etal-2022-nadi} & - / 18 / 5 & \\
         NADI2023 \cite{abdul-mageed-etal-2023-nadi} & - / 18 / 5 & - Currently not disclosed \\
         \midrule
         \textbf{(5) Miscellaneous} & & \\
         Arabic Dialects Dataset \cite{el-haj-etal-2018-arabic} & - / - / 4 * & - 12,801 sentences sampled from the AOC dataset, in addition to 3,693 sentences sampled from the \textit{Internet Forums} category of the Tunisian Arabic Corpus \cite{McNeil_Faiza_2010}.  \\

         \bottomrule
    \end{tabular}%
    }
    \caption{The list of single-labeled ADI datasets categorized by the labeling techniques. We follow the regional categorization of \citet{baimukan-etal-2022-hierarchical}. \textbf{Ct/Cn/Re}: the number of cities (provinces), countries, and regions respectively. \textbf{*}: The regional dialects are defined as Egypt, Iraq, Levant, Gulf, and Maghreb \cite{cotterell-callison-burch-2014-multi}. $\mathbf{\dag}$: Sudanese Arabic is considered as another regional dialect. \textbf{?}: Missing information.}
    \label{tab:ADI_datasets}
\end{table*}

\section{How are Current ADI Datasets Built?}

There have been multiple efforts to build several datasets for the ADI task using multiple techniques.
We recognize four main techniques: (1) Manual Human Annotation, (2) Translating sentences into predefined sets of dialects, (3) Automatic labeling of data using distinctive lexical cues, and (4) Automatic labeling using geo-tagging.

A common limitation to all those techniques is modeling the task as a single-label classification task, where each sentence in the datasets is assigned to only one dialect while ignoring the fact that the same sentence can be valid in multiple dialects. Furthermore, each of these techniques has its own additional limitations that affect the quality of the labels as follows:

\noindent \textbf{(1) Manual Human Annotation} where annotators categorize Arabic sentences into one dialect from a predefined list of dialects \cite{zaidan-callison-burch-2011-arabic, huang-2015-improved, malmasi-etal-2016-discriminating, zampieri-etal-2017-findings, zampieri-etal-2018-language}.
    
\noindent \underline{Limitations}: It was found that annotators over-identify their own native dialects \cite{zaidan-callison-burch-2014-arabic,abu-farha-magdy-2022-effect}. Therefore, the annotations for sentences that are valid in multiple dialects might be skewed toward the countries from which most of the annotators originate, causing a representation bias. Moreover, accurately determining the Arabic dialect of a sentence requires exposure to the different dialects of Arabic, which might not be a common case for Arabic speakers.\\

\begin{figure*}[!t]
    \centering
    \includegraphics[width=\textwidth]{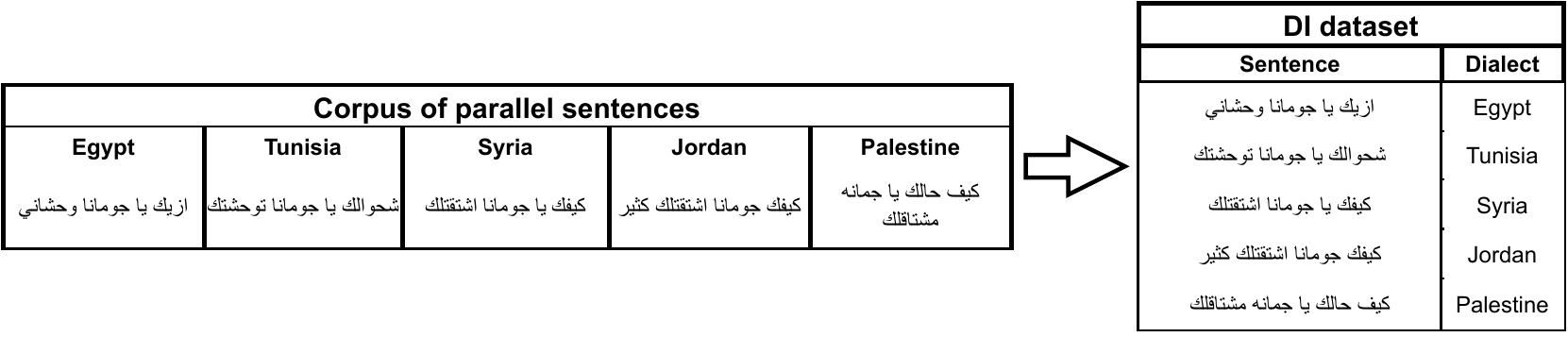}
    \caption{A demonstration of how parallel dialectal sentences are transformed into DI samples. The parallel sentences are sampled from the MPCA corpus \cite{bouamor-etal-2014-multidialectal}}
    \label{fig:parallel_to_DI}
\end{figure*}

\noindent \textbf{(2) Translation} in which participants are asked to translate sentences into their native Arabic dialects \cite{ho2016tatoeba, bouamor-etal-2014-multidialectal, meftouh-etal-2015-machine, bouamor-etal-2018-madar, MUBARAK18.13}. If all the participants are asked to translate the same source sentences, then the dataset is composed of parallel sentences in various dialects. The main application of these datasets is to help develop machine translation systems, however, they are sometimes used for ADI. Figure \ref{fig:parallel_to_DI} demonstrates how a corpus of parallel sentences is transformed into a corresponding DI dataset. 

\noindent \underline{Limitations}: While the labels of the corresponding DI dataset are correct, a source sentence might have the same translation in multiple Arabic dialects, Table \ref{tab:examples}. In such cases, a single-label classifier is asked to predict different \textit{Dialect} labels despite the input sentence being the same.

Moreover, the syntax, and lexical items in the translated sentences might be affected by the corresponding syntactic and lexical features of the source sentences, especially if the source sentence is MSA or a variant of DA \cite{bouamor-etal-2014-multidialectal,harrat:hal-01557405}. Such effects might make the translated sentences sound unnatural to native speakers of these dialects.\\

\noindent \textbf{(3) Distinctive Dialectal Terms} where text is curated based on the appearance of a term from a seed list of distinctive dialectal terms. These terms are used to automatically determine the dialect of the text \cite{alsarsour-etal-2018-dart,althobaiti_2022}.
    
\noindent \underline{Limitations}: The curated data is constrained by the diversity of the terms used to collect it.\\

\noindent \textbf{(4) Geo-tagging} where the text is automatically labeled using information about the location or the nationality of its writer \cite{mubarak-darwish-2014-using, salama-etal-2014-youdacc, al2016opinion, doi:10.1177/0165551516683908, zaghouani-charfi-2018-arap, charfi-etal-2019-fine, el-haj-2020-habibi, abdelali-etal-2021-qadi, abdul-mageed-etal-2020-nadi, abdul-mageed-etal-2021-nadi, abdul-mageed-etal-2022-nadi}.
    
\noindent \underline{Limitations}: While this technique allows for curating data from different Arab countries, it does not consider that speakers of a variant of DA might be living in an Arab country that speaks another variant (e.g.: An Egyptian living in Kuwait) \cite{charfi-etal-2019-fine, abdul-mageed-etal-2020-nadi}. Moreover, some of the curated sentences might be written in MSA, so the curated sentences need to be split into DA sentences and MSA ones \cite{abdelali-etal-2021-qadi, abdul-mageed-etal-2021-nadi, abdul-mageed-etal-2022-nadi}.

\section{Maximal Accuracy of Single-label ADI Datasets}
\label{sec:maximal_accuracy}
For a single-label ADI dataset consisting of sentences where each is assigned one dialect label, assume that a percentage $\mathbf{Perc_{2}}$ of those sentences is valid in $\mathbf{2}$ different dialects. For those sentences, only one of the valid dialects is listed as their label. An effective model trained to predict a single label will randomly assign each of these sentences to one of its two valid labels. Thus, the expected maximal accuracy on the dataset $\mathbf{E[Accuracy_{max}]}$ that the model can achieve would then be:
\begin{equation}
\label{eq:simplified}
    \small
        \mathbf{E[Accuracy_{max}] =\ }
        \mathbf{(100-Perc_{2}) + \frac{Perc_{2}}{2}}
\end{equation}

For example, if 40\% of the sentences are valid in two dialects (i.e.: $\mathbf{Perc_{2}} = 40\%$), then the $\mathbf{E[Accuracy_{max}]}$ of the dataset would be 80\%. 
This becomes worse when a sentence is valid in more dialects, exceeding ten valid dialects in some cases (as shown in Table \ref{tab:examples}). Thus, for a total number of dialects $N_{dialects}$, the equation above can then generalized to:

\begin{equation}
\label{eq:generic}
    \small
        \mathbf{E[Accuracy_{max}] =\ }
        \mathbf{Perc_{1}} + \mathbf{\sum_{n=2}^{n=N_{dialects}}{\frac{Perc_{n}}{n}}}
\end{equation}

where $\mathbf{Perc_{1}}$ is the percentage of samples that are only valid in one dialect, $\mathbf{Perc_{n}}$ is the percentage of samples valid in $n$ dialects, $N_{dialects}$ represent the total number of dialects considered, and $\mathbf{\sum_{n=1}^{n=N_{dialects}}{Perc_{n}} = 100\%}$.

The higher the percentages $\mathbf{Perc_{n}}$ where $n \in [2, N_{dialects}]$, the lower the maximal accuracy would be. The same pattern would apply to F1 scores. Therefore, a model might be achieving low F1 scores as a consequence of framing DI as a single-label classification task, which might result in high $\mathbf{Perc_{n}}$ values.

Our objective in this paper is to estimate the value of $\mathbf{E[Accuracy_{max}]}$ for the existing datasets, which should examine the validity of our hypothesis that modeling ADI task as a single-label classification can be highly sub-optimal.

\section{Estimating the Maximal Accuracy of Datasets}
\label{sec:estimation_percentage}

\begin{table*}[!t]
    \centering
    \small
    \begin{tabular}{lcccc}
         \textbf{Dataset} & $\mathbf{N_{dialects}}$ & $\mathbf{N_{samples}}$ & $\mathbf{\sum_{n=2}^{n=N_{dialects}}{\widetilde{P}erc_{n}}}$ & $\mathbf{\widetilde{E}[Accuracy_{max}]}$ \\
         \midrule
         \textbf{PADIC} & 4 & 29,138 & 5.2\% & 97.1\%\\
         \textbf{MPCA} & 5 & 4,960 & 7.8\% & 95.4\%\\
         \textbf{MADAR6} & 5 & 49,476 & 2.3\% & 98.7\%\\
         \textbf{MADAR26} & 15 & 48,624 & 9.6\% & 93.9\%\\
         \bottomrule
    \end{tabular}
    \caption{The estimated percentages and the corresponding expected maximal accuracy for the DI datasets formed using the four parallel corpora. The estimated maximal accuracies are upper bounds for the true maximal accuracies, and we expect the true values to be significantly lower than these estimates.}
    \label{tab:parallel_corpora_stats}
\end{table*}

\begin{figure*}[!t]
    \centering
    \includegraphics{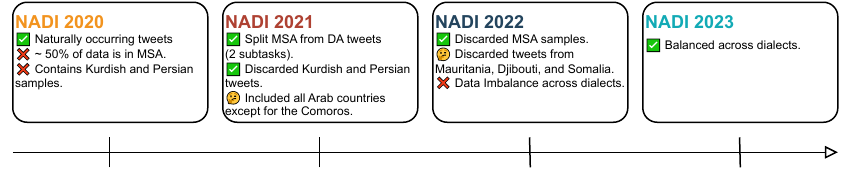}
    \caption{The evolution of the NADI datasets used for the shared tasks run between 2020 and 2023.}
    \label{fig:NADI_shared_tasks}
\end{figure*}

In our study, we focus on the country-level ADI for which multiple shared tasks have been organized since 2019 \cite{bouamor-etal-2019-madar,abdul-mageed-etal-2020-nadi,abdul-mageed-etal-2021-nadi,abdul-mageed-etal-2022-nadi}.

In order to quantify the percentages $Perc_{n}$, each sample of a dataset needs to be assessed by native speakers from all the Arab countries. Given our inability to recruit participants from all the Arab countries, we will estimate the percentages using two methods that provide lower bounds $\mathbf{\widetilde{P}erc_{n}}$ for the actual values $\mathbf{Perc_{n}}$ (i.e.: $\mathbf{\widetilde{P}erc_{n}} \le \mathbf{Perc_{n}}$). Consequently, the estimated maximal accuracy is an upper bound for its true value.

\subsection{Datasets Derived from Parallel Corpora}
Initially, we examine the possibility of having Arabic sentences valid in multiple dialects by examining parallel corpora of Arabic dialects, which have sentences translated into multiple dialects. While a manual translation of a sentence can be phrased in different forms within the same dialect, we still examine if by chance we can find identical manually-translated sentences in different dialects by different translators.

For the four parallel corpora \textbf{Multidialectal Parallel Corpus of Arabic (MPCA)} \cite{bouamor-etal-2014-multidialectal}, \textbf{PADIC} \cite{meftouh-etal-2015-machine}, \textbf{MADAR6}, and \textbf{MADAR26} \cite{bouamor-etal-2018-madar}, we transformed the parallel sentences into \textit{(sentence, dialect)} pairs as in subtask (1) of the MADAR shared task \cite{bouamor-etal-2019-madar}. We then mapped the dialect labels for \textbf{PADIC}, \textbf{MADAR6}, and \textbf{MADAR26} from city-level dialects to country-level ones. In case the same sentence is used in different cities within the same country, a single copy is kept. The sentences are then preprocessed by discarding Latin and numeric characters in addition to diacritics and punctuation. Lastly, we estimated the percentages $\mathbf{\widetilde{P}erc_{n}}$ by computing the percentages of sentences that have the exact same translation in $n$ dialects.

The upper bound for the maximal accuracies of the four corpora lies in the range $[93.9\%, 98.7\%]$ as per Table \ref{tab:parallel_corpora_stats}. The fact that the maximal accuracy for \textbf{MADAR26} is lower than that for \textbf{MADAR6} demonstrates that the probability that a sentence is valid in multiple dialects increases as more translations in other country-level dialects are considered.

\subsection{Datasets of Geolocated Dialectal Sentences}
\label{sec:geolocated_datasets}

The Nuanced Arabic Dialect Identification (NADI) shared tasks \cite{abdul-mageed-etal-2020-nadi,abdul-mageed-etal-2021-nadi,abdul-mageed-etal-2022-nadi} used datasets that are built by collecting Arabic tweets authored by users who have been tweeting from the same location for 10 consecutive months. The geolocation of the users is then used as a label for their tweets. The creators of NADI have been improving the quality of the dataset from one year to another as summarized in Figure \ref{fig:NADI_shared_tasks}. 

While the NADI shared tasks have been attracting active participation, the best-performing models in NADI 2022 achieved macro F1 scores of 36.48\% and 18.95\%, and accuracies of 53.05\% and 36.84\% on two test sets \cite{abdul-mageed-etal-2022-nadi}. The baseline MarBERT-based model \cite{abdul-mageed-etal-2021-arbert} fine-tuned on the training dataset achieves competitive results (macro F1 scores: 31.39\% and 16.94\%, accuracies: 47.77\% and 34.06\%).

\paragraph{Model Description} Given the competitiveness of the baseline model, we fine-tuned the MarBERT model on the balanced training dataset of NADI 2023, and then we used the QADI dataset \cite{abdelali-etal-2021-qadi} as our test set. QADI's test set covers the same 18 countries as NADI 2023. 
We decided to analyze the errors of our model on QADI for two reasons: 1) At the time of writing the paper, the test set of NADI 2023 was not released (even for earlier NADIs, the labels of the test sets are not publicly released); 2) The dialect labels of the samples of QADI's test set were automatically assigned using geolocations similar to NADI, but the label of each sample was validated by a native speaker of the sample's label, which gives additional quality assurance for QADI over NADI. 

\begin{figure}[!t]
    \centering
        \centering
        \includegraphics[width=\columnwidth]{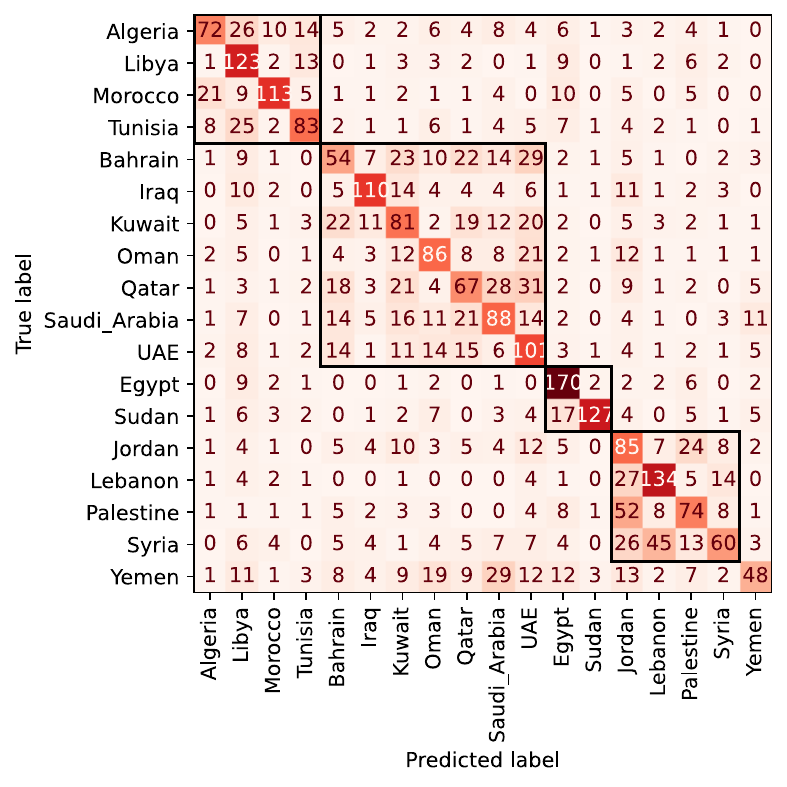}
        \caption{The confusion matrix for the predictions of a MarBERT model on QADI's test set. The model was fine-tuned using NADI 2023's training dataset.}
        \label{fig:confusion_matix}
\end{figure}

The model achieves an accuracy of \textbf{50.74\%} on QADI's test set with the full classification report in Table \ref{tab:QADI_metrics}. Figure \ref{fig:confusion_matix} visualizes how the predictions and labels are confused together.

\paragraph{Manual Error Analysis} We recruited native-speaker participants of Algerian, Egyptian, Palestinian, Lebanese, Saudi Arabian, Sudanese, and Syrian Arabic to validate the False Positives (FPs) that the model makes for those dialects. Each participant is shown the FPs for their native dialect, one at a time, and is asked to validate them as indicated in Figure \ref{fig:qualtrics_screenshots} \footnote{We release the judgments through: \url{github.com/AMR-KELEG/ADI-under-scrutiny/tree/master/data}}.
If the participant found the FP sample to be valid in their native dialect, it means that this sample is valid in at least two different Arabic dialects (i.e.: the sample's original label, and the model's prediction) \footnote{Participants are given a third choice \textit{Maybe / Not Sure}, which we count as \textit{No} (i.e.: invalid in their dialect).}. However, it can still be valid in additional dialects, which we did not check for due to the limited number of participants.

\begin{figure}[!t]
    \centering
    \includegraphics{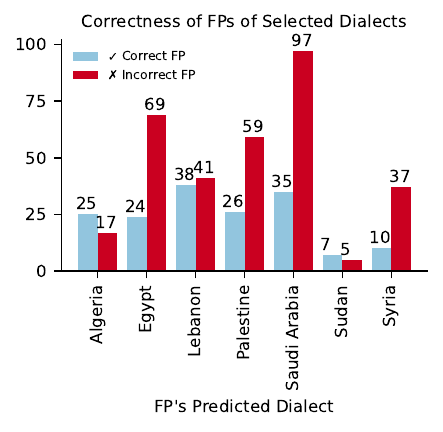}
    \caption{The distribution of the annotations for the validity of the False Positives (FPs) in 7 Arabic dialects.\\\textbf{\textcolor{correct_FP}{Correct FP}} represents the FP samples for which the model's prediction is invalid. \textbf{\textcolor{incorrect_FP}{Incorrect FP}} the FP samples for which the model's prediction is valid.}
    \label{fig:FP_distribution}
\end{figure}

\begin{figure*}[!t]
    \centering
    \includegraphics[width=0.85\textwidth]{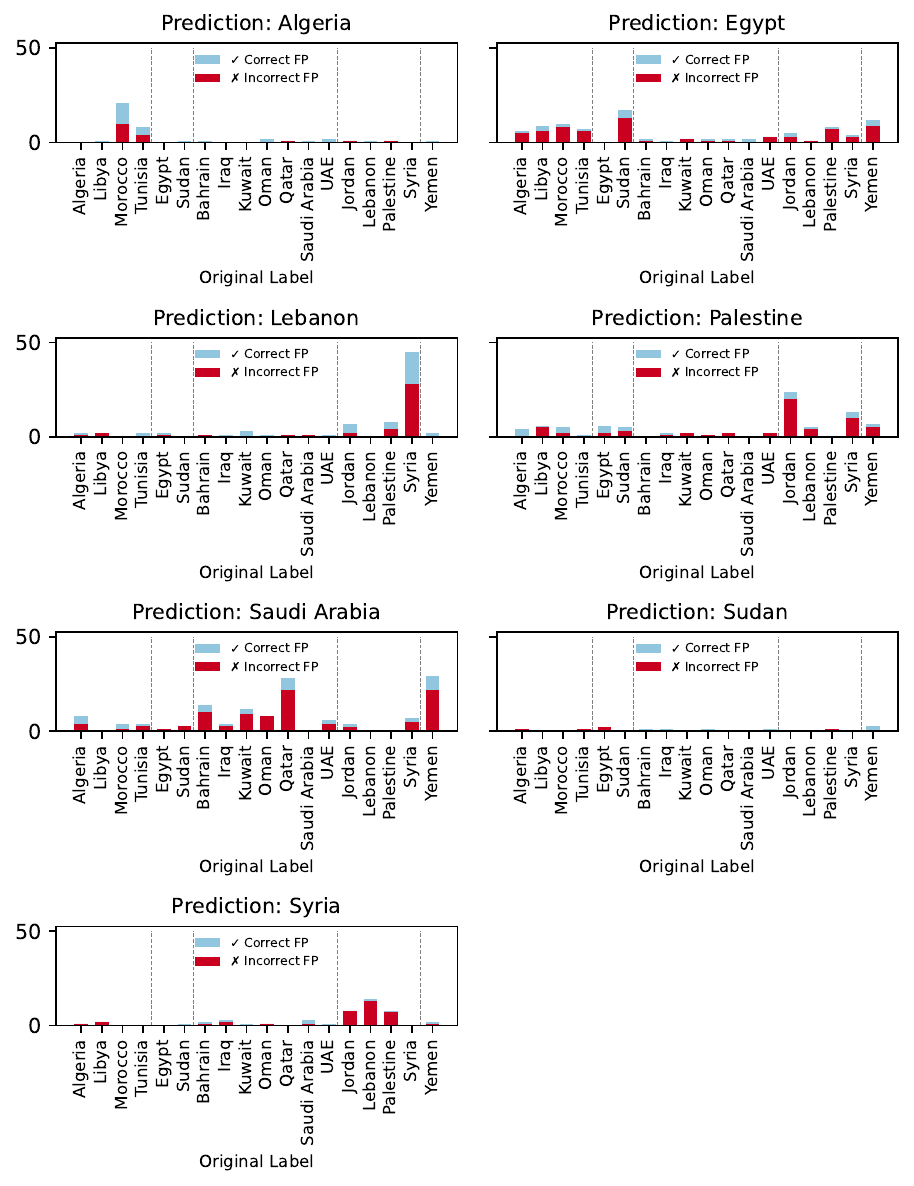}
    \caption{The distribution of the original labels for the False Positives (FPs) of the seven validated dialects.\\ \textbf{\textcolor{correct_FP}{Correct FP}} represents the FP samples for which the model's prediction is invalid. \textbf{\textcolor{incorrect_FP}{Incorrect FP}} represents the FP samples for which the model's prediction is valid.}
    \label{fig:detailed_FP_distribution}
\end{figure*}

\paragraph{Validity of the Model's FPs} Out of 490 validated FPs, 325 were found to be also valid in the other dialect they were classified to, which represents $\approx 66\%$ of the validated errors. Having such a great proportion of FPs that are not true errors hinders the ability to properly analyze and improve the ADI models. For Egyptian, Palestinian, Saudi Arabian, and Syrian Arabic, the majority of the FPs are incorrect as demonstrated in Figure \ref{fig:FP_distribution} (i.e.: the model's prediction should be considered to be correct). As expected, dialects grouped in the same region are similar, and thus the FPs of a dialect would generally have labels of other dialects from the same region as in Figure \ref{fig:detailed_FP_distribution}.

\paragraph{Impact on Evaluation} If we only consider the 725 samples that were correctly predicted by the model (TPs) in addition to the validated 490 FPs, then we know that 325 samples out of 1215 ones are at least valid in two different dialects. The $\mathbf{\tilde{P}erc_{2}}$ for this subset is $26.7\%$, making the maximal accuracy $\mathbf{E[Accuracy_{max}]}$ equal to $86.6\%$.

To further investigate the impact of the incorrect FPs on the evaluation metrics, we computed the corrected True Positive value for each dialect $\mathbf{TP^{*}}$ as $\mathbf{TP^{*} = TP + Incorrect\ FP}$. Using these corrected $\mathbf{TP^{*}}$ values, we computed corrected precision, recall, and F1-scores. As per Table \ref{tab:updated_scores}, the macro-averaged F1-score increased from 0.56 to 0.72. This clearly confirms our hypothesis that modeling ADI task as a single-label classification task leads to inaccurate evaluation of the systems.

\begin{table*}[t]
    \aboverulesep=0ex 
    \belowrulesep=0ex 
    \rule{0pt}{1.1EM}
    \centering
    \small

    \begin{tabular}{c|ccccc|ccc|ccc}
     \textbf{Dialect} & $\mathbf{TP}$ & $\mathbf{FP}$ & $\mathbf{TP^{*}}$ & $\mathbf{FP^{*}}$ & $\mathbf{FN}$ & $\mathbf{P}$ &  $\mathbf{R}$ &  $\mathbf{F1}$ & $\mathbf{P^{*}}$ & $\mathbf{R^{*}}$ & $\mathbf{F1^{*}}$ \\

    \midrule
    \rule{0pt}{1.1EM}
    Algeria &  72 &  42 \tikzmark{s} &   72 + \tikzmark{t} 17 = 89 &         \tikzmark{u} 25 & 98 &      0.63 &    0.42 &      0.51 &               0.78 &              0.48 &              0.59\\
    &&&&&&&&&&&\\
    Egypt & 170 &  93 \tikzmark{p} & 170 + \tikzmark{q} 69 = 239 &         \tikzmark{r} 24 &  30 &     0.65 &    0.85 &      0.73 &               0.91 &             0.89 &              0.90 \\
    &&&&&&&&&&&\\
    Lebanon & 134 &  79 \tikzmark{m} & 134 + \tikzmark{n} 41 = 175 &         \tikzmark{o} 38 & 60 &      0.63 &    0.69 &      0.66 &               0.82 &              0.74 &              0.78 \\
    &&&&&&&&&&&\\
    Palestine &  74 &  85 \tikzmark{j} &  74 + \tikzmark{k} 59 = 133 &         \tikzmark{l} 26 & 99 &      0.47 &    0.43 &      0.45 &               0.84 &              0.57 &              0.68 \\
    &&&&&&&&&&&\\
    Saudi\ Arabia &  88 & 132 \tikzmark{g} &  88 + \tikzmark{h} 97 = 185 &         \tikzmark{i} 35 &  111 &     0.40 &    0.44 &      0.42 &               0.84 & 0.62 &              0.72 \\
    &&&&&&&&&&&\\
       Sudan & 127 &  12 \tikzmark{d} &  127 + \tikzmark{e} 5 = 132 &          \tikzmark{f} 7 & 61 &      0.91 &    0.68 &      0.78 &               0.95 & 0.68 &              0.80 \\
    &&&&&&&&&&&\\
       Syria &  60 &   47 \tikzmark{a} &   60 + \tikzmark{b} 37 = 97 &         \tikzmark{c} 10 & 134 &      0.56 &    0.31 &      0.40 &               0.91 & 0.42 &              0.57 \\
    &&&&&&&&&&&\\
    \midrule
    \rule{0pt}{1.1EM}

    \textbf{Macro-average} & & & & & & 0.61 & 0.55 & 0.56 & 0.86 & 0.63 &              0.72 \\
    \bottomrule
    \end{tabular}
    \caption{The impact of the incorrect FPs on the precision $\mathbf{P}$, recall $\mathbf{R}$, and F1-score $\mathbf{F1}$. Error samples for a specific predicted dialect (i.e.: FPs of this dialect) that are labeled as valid in this predicted dialect are counted as true positives in the corrected $\mathbf{TP^{*}}$ score. The corrected $\mathbf{P^{*}}$, $\mathbf{R^{*}}$ and $\mathbf{F1^{*}}$ are based on the corrected value of $\mathbf{TP^{*}}$.\\
    $\mathbf{P^{*} = \frac{TP^{*}}{TP^{*}+FP^{*}}}$,     $\mathbf{R^{*} = \frac{TP^{*}}{TP^{*}+FN}}$,     $\mathbf{F1^{*} = \frac{2 * P^{*} * R^{*}}{P^{*} + R^{*}}}$\\
    \textbf{Note:} $P$ stands for Precision, $R$ stands for Recall, and $F1$ stands for F1-score.
    }
    \label{tab:updated_scores}
    \begin{tikzpicture}[overlay, remember picture, shorten >=.5pt, shorten <=.5pt, transform canvas={yshift=.25\baselineskip}]
    \draw [->] ({pic cs:a}) [bend right] to ([xshift=2pt, yshift=-3pt]{pic cs:b});
    \draw [->] ({pic cs:a}) [bend right] to ({pic cs:c});
    
    \draw [->] ({pic cs:d}) [bend right] to ([xshift=2pt, yshift=-3pt]{pic cs:e});
    \draw [->] ({pic cs:d}) [bend right] to ({pic cs:f});

    \draw [->] ({pic cs:g}) [bend right] to ([xshift=2pt, yshift=-3pt]{pic cs:h});
    \draw [->] ({pic cs:g}) [bend right] to ({pic cs:i});

    \draw [->] ({pic cs:j}) [bend right] to ([xshift=2pt, yshift=-3pt]{pic cs:k});
    \draw [->] ({pic cs:j}) [bend right] to ({pic cs:l});

    \draw [->] ({pic cs:m}) [bend right] to ([xshift=2pt, yshift=-3pt]{pic cs:n});
    \draw [->] ({pic cs:m}) [bend right] to ({pic cs:o});

    \draw [->] ({pic cs:p}) [bend right] to ([xshift=2pt, yshift=-3pt]{pic cs:q});
    \draw [->] ({pic cs:p}) [bend right] to ({pic cs:r});

    \draw [->] ({pic cs:s}) [bend right] to ([xshift=2pt, yshift=-3pt]{pic cs:t});
    \draw [->] ({pic cs:s}) [bend right] to ({pic cs:u});
  \end{tikzpicture}
\end{table*}


\section{Proposal for Framing the ADI Task}
\label{sec:multi-label}

Given the limitations of using single-label classification for the ADI task, elaborated in \S\ref{sec:estimation_percentage}, we propose alternative modeling for the task. 

\citet{zaidan-callison-burch-2014-arabic} asked crowdsourced annotators to label dialectal sentences as being \textit{Egyptian}, \textit{Gulf}, \textit{Iraqi}, \textit{Levantine}, \textit{Maghrebi}, \textit{other dialect}, or \textit{general dialect}. They used the \textit{general dialect} for sentences that can be valid in multiple dialects. The \textit{general dialect} is underspecified, and it is not clear whether it implies that a sentence is accepted in multiple dialects or in all of them. Therefore, the authors noticed that some of the annotators barely used the label, while others used it when they were not sure about the dialect of the underlying sentences. Moreover, they noticed that the annotators tend to over-identify their native dialects. Annotators might not realize that a sentence valid in their native dialect is also valid in other dialects, and thus can end up choosing their native dialect as the label for this sentence, instead of the \textit{general dialect} label.

\citet{zampieri2023language} focused on the binary distinction between two varieties of English, Portuguese, and Spanish. In addition to the two varieties of each language, the annotators are allowed to assign sentences to a third label \textit{Both or Neither}. The evaluation results indicate that the \textit{Both or Neither} label is harder to model computationally than the other variety labels. The authors noted that there is room for improvement in the treatment and modeling of this third label.

Consequently, we believe that adding another label such as \textit{general} or \textit{Both or Neither} does not completely solve the limitations of single-label classification datasets. Conversely, framing the task as a multi-label classification would optimally alleviate the aforementioned limitations.

\subsection{ADI as Multi-label Classification}

Multi-label classification allows assigning one or more dialects to the same sample. \citet{bernier-colborne-etal-2023-dialect} argued for using the multi-label classification setup after investigating a French DI corpus (FreCDo) \cite{gaman2022frecdo}, covering four macro French dialects spoken in France, Switzerland, Belgium, and Canada. They found that the corpus has duplicated single-labeled sentences of different labels, and showed how these sentences impact the performance of DI models.

\textbf{Labeling}: Collecting multi-labels for a dataset requires the manual annotation of its samples. Dataset creators need to consider how they collect the annotations, and consequently who to recruit. An Arabic speaker of a specific dialect would be able to determine if a sentence is valid in their dialect or not \cite{salama-etal-2014-youdacc, abdelali-etal-2021-qadi}. \citet{althobaiti_2022} found that the average inter-annotator agreement score (Cohen's Kappa) is 0.64, where two native speakers of 15 different country-level Arabic dialects are asked to check the validity of tweets in their native dialects.

While human participants can sometimes infer the macro-dialect of a sentence that is not in their native dialect, it seems quite hard for them to predict the country-level dialects in which the sentence is valid \cite{abdul-mageed-etal-2020-toward}.

\noindent\underline{Recommendation}: Ask Arabic speakers to identify if a sentence is valid in their native dialects or not as per \cite{salama-etal-2014-youdacc, abdelali-etal-2021-qadi, althobaiti_2022}. In order to include new dialects, speakers of these dialects need to be recruited. 

\textbf{Modeling}: One way of building multi-label classification models is to use multiple binary classifiers. More specifically, a binary classifier is built to decide whether a sentence is valid in one dialect or not. For $N$ dialects, $N$ binary classifiers would be responsible for predicting the labels of a single sample.

\textbf{Evaluation}: For each supported dialect, evaluation metrics like accuracy, precision, recall, and F1-score can be used. Macro-averaging the metrics is a way to measure the average performance of the model across the different dialects. 

\textbf{Extensibility}: The multi-label framing is extensible since more labels can be added to a previously annotated dataset. Adding a new dialect class does not invalidate the labels of the other dialect classes. 

This does not apply to the single-label framing since an annotator would need to select a dialect out of a predefined set of dialects. Changing the set of dialects would require the reannotation of the whole dataset.

\section{Conclusion}
\label{sec:conclusion}
Single-label classification has been the defacto framing for Arabic Dialect Identification (ADI). We show that such framing implies that any model would have a maximal accuracy that is less than 100\%, since some samples are valid in multiple dialects, and thus their labels are randomly assigned from these dialects in which they are valid. For a set of 490 validated False Positives (FPs) of an ADI model, we found that the model's predicted dialects for 325 of them are also valid. The fact that about 66\% of the FPs are not true errors hinders the ability to analyze and improve the ADI models, and hurts the reliability of the evaluation metrics.

Given this major limitation of single-label framing, we argue that ADI should be framed as a multi-label task. This follows the recommendation of \citet{bernier-colborne-etal-2023-dialect} for French Dialect Identification. We hope that this paper will spark discussions across the Arabic NLP community about the current state of ADI, and encourage the creation of new datasets in a multi-label setup, with labels assigned manually by native speakers of the different Arabic dialects.

For future work, we will investigate the impact of the Arabic Level of Dialectness (ALDi) variable introduced by \citet{ALDi} on identifying the dialect of sentences. Intuitively, the dialect of a sentence with a high ALDi score is easier to identify since the sentence shows more features of dialectness than those of sentences having low ALDi scores. Therefore ALDi can be used to identify the samples that are more expected to be valid in multiple dialects, facilitating the annotation process of new DI datasets.

\section*{Limitations}
Recruiting native speakers from the 18 Arab countries included in the NADI 2023 dataset proved to be hard. Moreover, we opted to only annotate the sentences of QADI's test set that were misclassified by the model. In order to accurately estimate the maximal accuracy for a dataset, all the samples should be checked independently by native speakers of the 18 supported Arab countries.

\section*{Acknowledgments}
This work was supported by the UKRI Centre for Doctoral Training in Natural Language Processing, funded by the UKRI (grant EP/S022481/1) and the University of Edinburgh, School of Informatics.

The error analysis experiment, in which we asked human participants to identify if sentences are valid in their native dialects, was approved by the research ethics committee of the University of Edinburgh School of Informatics with reference number 207712.

\bibliography{anthology,custom}
\bibliographystyle{acl_natbib}

\appendix
\setcounter{table}{0}
\setcounter{figure}{0}
\renewcommand{\thetable}{\Alph{section}\arabic{table}}
\renewcommand{\thefigure}{\Alph{section}\arabic{figure}}

\section{Detailed Dialect Coverage and Model Performance Report}
\begin{table*}[!t]
    \centering
    \small
    \begin{tabular}{lp{5.5cm}p{5.5cm}}
         \textbf{Dataset} & \multicolumn{1}{c}{\textbf{Cities}} & \multicolumn{1}{c}{\textbf{Countries}} \\
         \midrule
         \multirow{2}{*}{\textbf{PADIC}} & \multicolumn{1}{c}{N = 5} & \multicolumn{1}{c}{N = 4} \\
         & Annaba, Algiers, Sfax, Damascus, Gaza & Algeria, Tunisia, Syria, Palestine \\
         \midrule
         \multirow{2}{*}{\textbf{MPCA}} & \multicolumn{1}{c}{\multirow{2}{*}{N/A}} & \multicolumn{1}{c}{N = 5} \\
         & & Egypt, Syria, Jordan, Palestine, Tunisia \\
         \midrule
         \multirow{2}{*}{\textbf{MADAR6}} & \multicolumn{1}{c}{N = 5} & \multicolumn{1}{c}{N = 5} \\
         & Beirut, Cairo, Doha, Tunis, Rabat & Lebanon, Egypt, Qatar, Tunisia, Morocco \\
         \midrule
         \multirow{2}{*}{\textbf{MADAR26}} & \multicolumn{1}{c}{N = 25} & \multicolumn{1}{c}{N = 15} \\
         & Aleppo, Damascus, Algiers, Alexandria, Aswan, Cairo, Amman, Salt, Baghdad, Basra, Mosul, Beirut, Benghazi, Tripoli, Doha, Fes, Rabat, Jeddah, Riyadh, Jerusalem, Khartoum, Muscat, Sanaa, Sfax, Tunis
         & Syria, Algeria, Egypt, Jordan, Iraq, Lebanon, Libya, Qatar, Morocco, Saudi Arabia, Palestine, Sudan, Oman, Yemen, Tunisia\\
         \midrule
         \textbf{QADI} & \multicolumn{1}{c}{\multirow{2}{*}{N/A}} & \multicolumn{1}{c}{N = 18} \\
         \textbf{NADI 2023} & & Algeria, Bahrain, Egypt, Iraq, Jordan, Kuwait, Lebanon, Libya, Morocco, Oman, Palestine, Qatar, Saudi Arabic, Sudan, Syria, Tunisia, United Arab Emirates, Yemen\\
         \bottomrule
    \end{tabular}
    \caption{The list of labels in the different ADI datasets.}
    \label{tab:datasets_classes}
\end{table*}

\begin{table*}[t]
    \aboverulesep=0ex 
    \belowrulesep=0ex 
    \centering
    \begin{tabular}{lc|ccc}
        \textbf{Dialect} &  \textbf{Support} & \textbf{Precision (P)} &  \textbf{Recall (R)} &  \textbf{F1-score (F1)} \\
        \midrule
        Algeria      &      170 &       0.63 &    0.42 &      0.51 \\
        Libya        &      169 &       0.45 &    0.73 &      0.56 \\
        Morocco      &      178 &       0.77 &    0.63 &      0.70 \\
        Tunisia      &      154 &       0.63 &    0.54 &      0.58 \\
        Bahrain      &      184 &       0.33 &    0.29 &      0.31 \\
        Iraq         &      178 &       0.69 &    0.62 &      0.65 \\
        Kuwait       &      190 &       0.38 &    0.43 &      0.40 \\
        Oman         &      169 &       0.46 &    0.51 &      0.49 \\
        Qatar        &      198 &       0.37 &    0.34 &      0.35 \\
        Saudi\ Arabia &      199 &       0.40 &    0.44 &      0.42 \\
        UAE          &      192 &       0.37 &    0.53 &      0.43 \\
        Egypt        &      200 &       0.65 &    0.85 &      0.73 \\
        Sudan        &      188 &       0.91 &    0.68 &      0.78 \\
        Jordan       &      180 &       0.31 &    0.47 &      0.38 \\
        Lebanon      &      194 &       0.63 &    0.69 &      0.66 \\
        Palestine    &      173 &       0.47 &    0.43 &      0.45 \\
        Syria        &      194 &       0.56 &    0.31 &      0.40 \\
        Yemen        &      193 &       0.55 &    0.25 &      0.34 \\
        \midrule        
        \multicolumn{2}{c}{\textbf{Macro avg.}}     &    0.5309 &  0.5085 &    0.5072 \\
        \multicolumn{2}{c}{\textbf{Weighted avg.}}   &     0.5295 &  0.5074 &    0.5058 \\
        \midrule
        \multicolumn{2}{c}{\textbf{Accuracy}}       &    0.5074 & &  \\
        \bottomrule
    \end{tabular}
    \caption{The evaluation metrics for the predictions of the fine-tuned MarBERT model on QADI's testing set. The model is fine-tuned on NADI 2023's training data.}
    \label{tab:QADI_metrics}
\end{table*}

The datasets used in the paper cover different Arabic dialects as detailed in Table \ref{tab:datasets_classes}. The \textbf{PADIC} dataset covers 4 country-level Arabic dialects from North Africa (Algeria, Tunisia), and the Levant (Syria, Palestine). On the other hand, the \textbf{QADI}, and \textbf{NADI 2023} datasets cover 18 country-level Arabic dialects.

Covering more dialects in a dataset impacts the performance of ADI models. Table \ref{tab:QADI_metrics} provides the detailed performance report of the MarBERT model fine-tuned for ADI between 18 country-level dialects, using NADI 2023's training dataset.

\section{The Error Analysis Survey}
\begin{figure*}[!ht]
     \centering
    \begin{subfigure}[b]{0.45\textwidth}
         \centering
         \fbox{\includegraphics[trim={0 4cm 0 0}, clip, width=\textwidth]{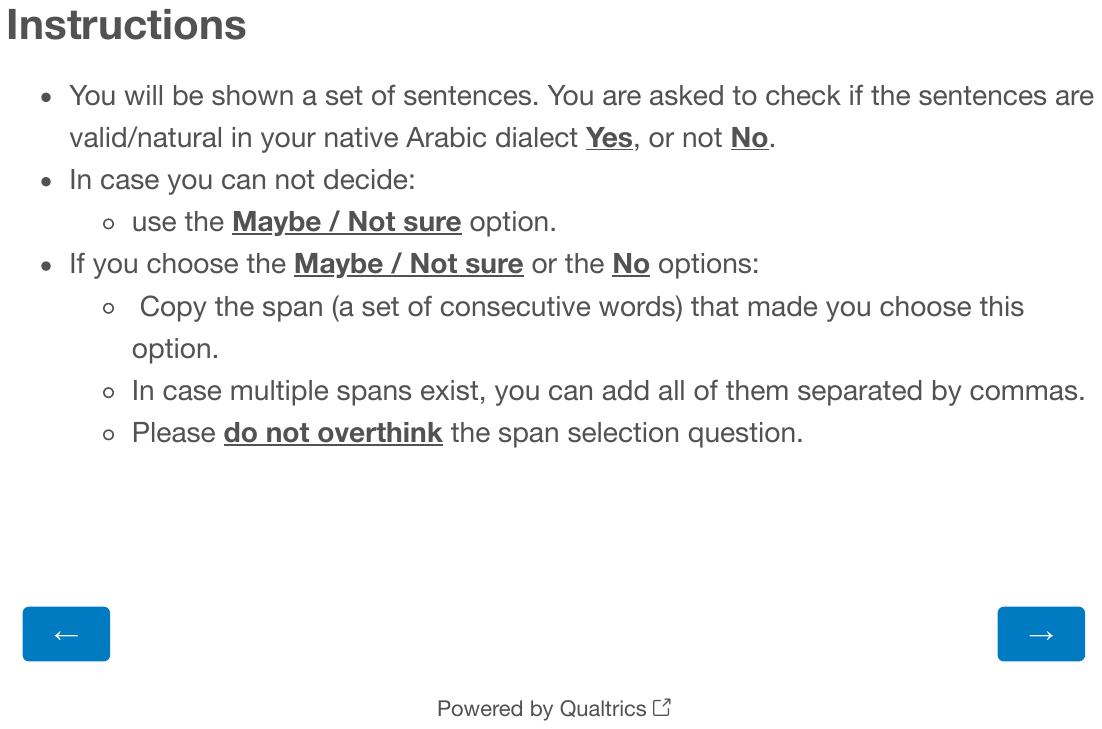}}
         \caption{Instructions page.}
    \end{subfigure}
    \hfill
    \begin{subfigure}[b]{0.45\textwidth}
         \centering
         \fbox{\includegraphics[trim={0 4cm 0 0}, clip, width=\textwidth]{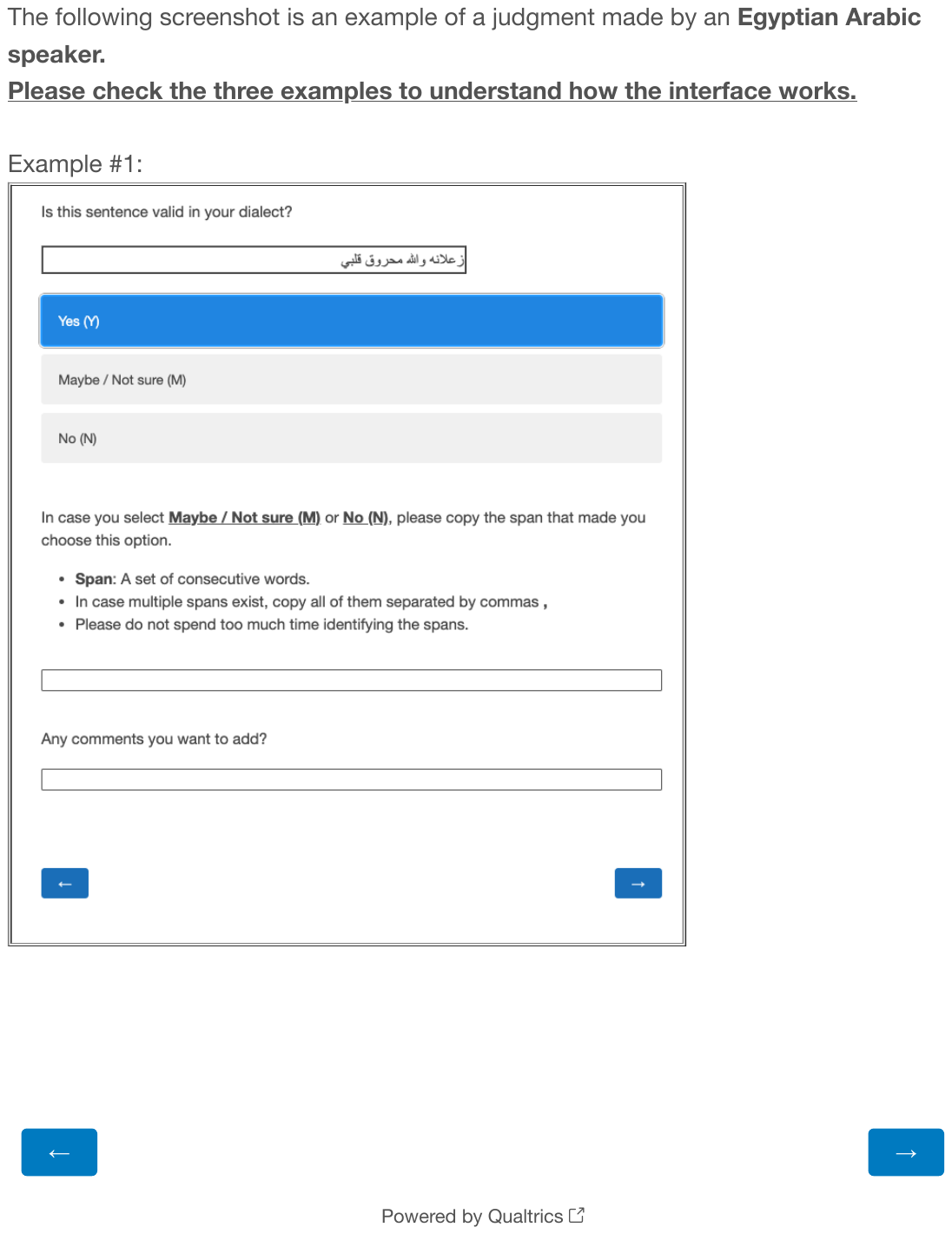}}
         \caption{First example page.}
    \end{subfigure}

    \begin{subfigure}[b]{0.45\textwidth}
         \centering
         \fbox{\includegraphics[trim={0 5cm 0 0}, clip, width=\textwidth]{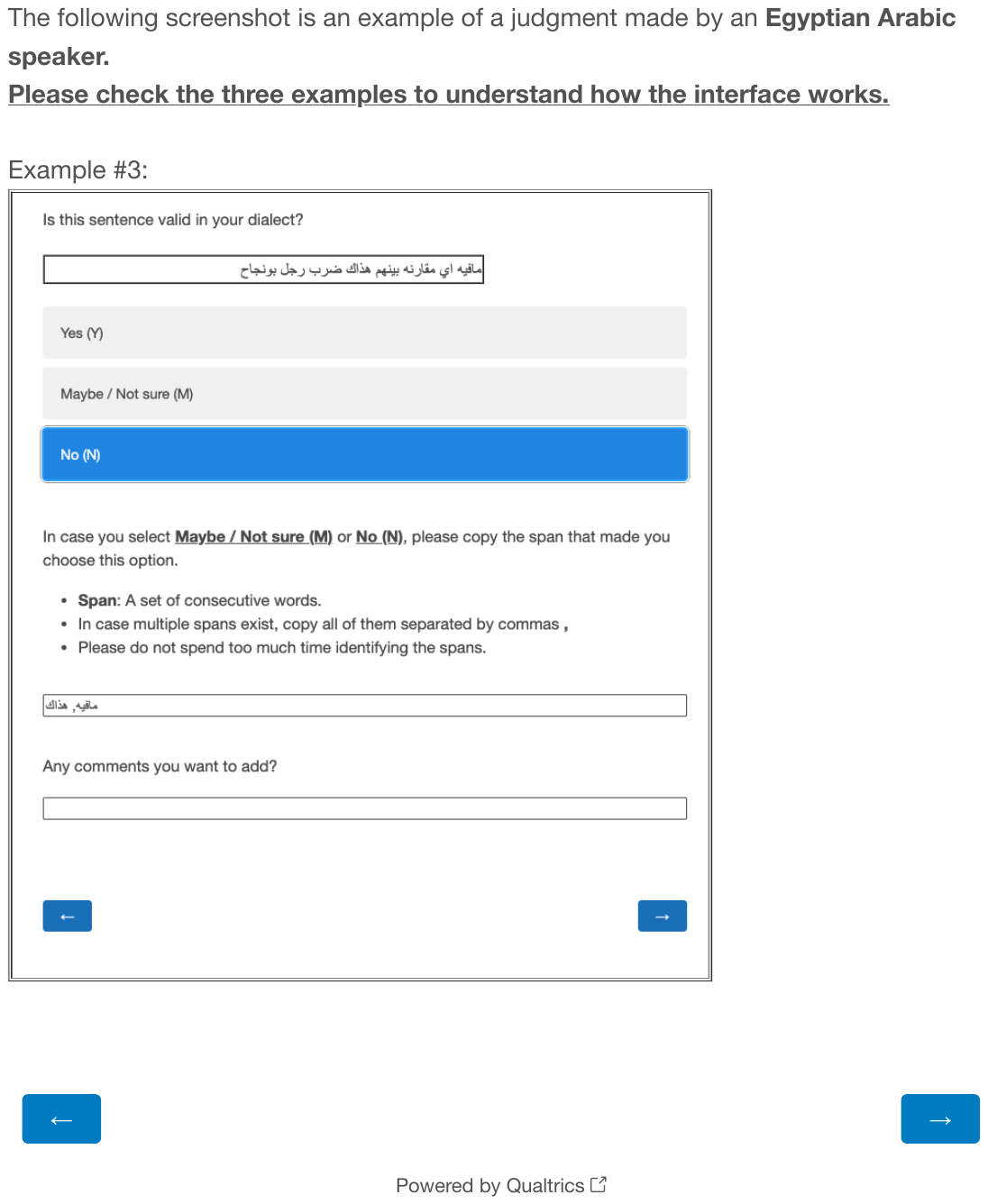}}
         \caption{Third example page.}
     \end{subfigure}
     \hfill
     \begin{subfigure}[b]{0.45\textwidth}
         \centering
         \fbox{\includegraphics[trim={0 5cm 0 0}, clip, width=\textwidth]{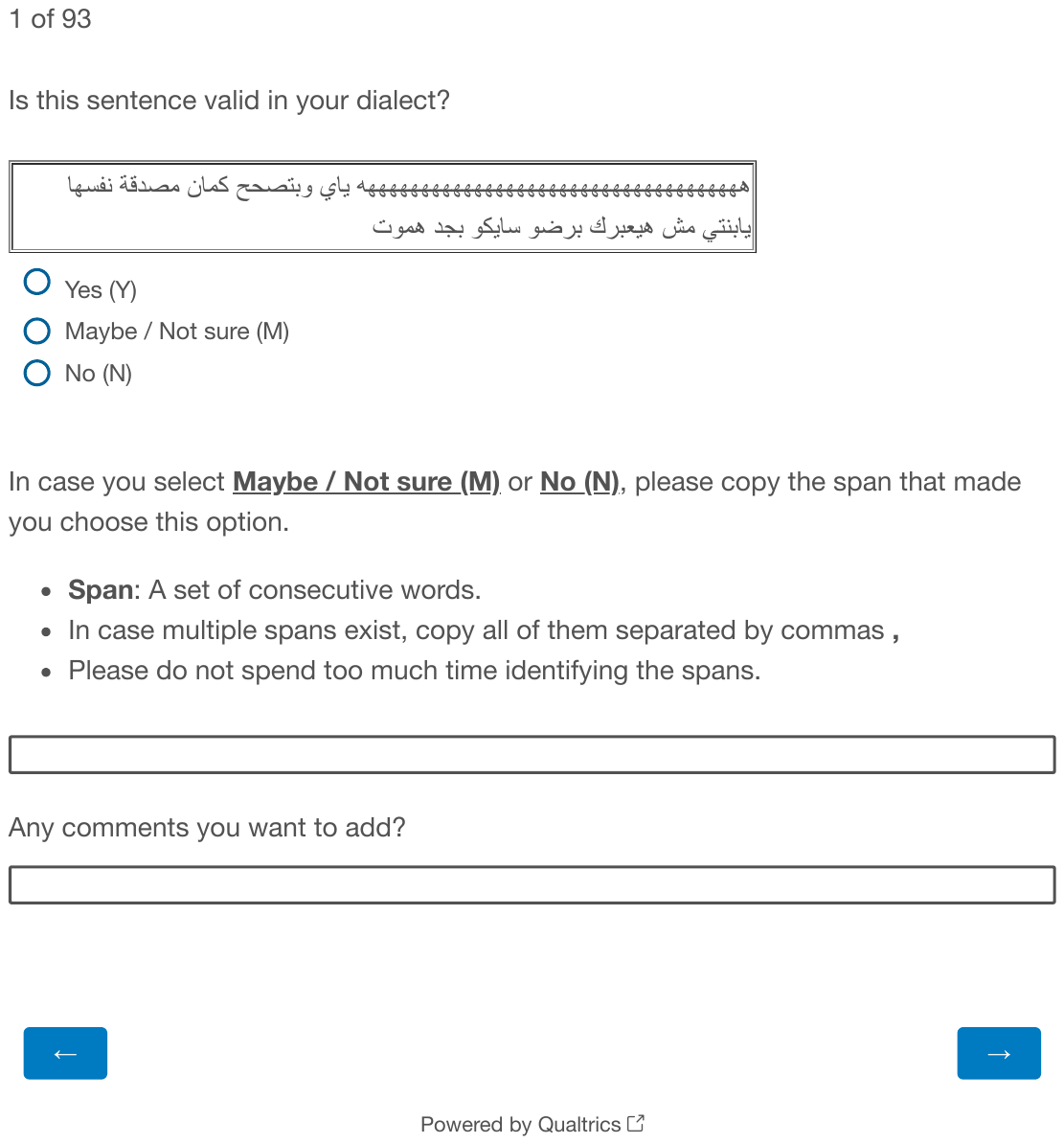}}
         \caption{An annotation page.}
     \end{subfigure}
     \caption{Screenshots of the different pages of the annotation task described in \S\ref{sec:geolocated_datasets}.}
     \label{fig:qualtrics_screenshots}
\end{figure*}

We created an online survey to validate the False Positives (FPs) of the MarBERT model fine-tuned on NADI 2023's training dataset. The survey aims to validate whether the errors of the model are caused by the single-label limitation of the testing dataset or are actual errors. Figure \ref{fig:qualtrics_screenshots} shows screenshots of the Instructions, Sample examples, and the annotation interface.

Table \ref{tab:valid_FPs} lists some examples for samples of the QADI dataset for which the model's predictions do not match the original labels, yet the annotators found these predictions to  also be valid.

\begin{table*}[!ht]
    \centering
    \begin{tabular}{lrc}
         \textbf{Valid Label} & \textbf{Sentence} & \textbf{Original Label} \\
         \midrule
         \multirow{2}{*}{\textbf{Algeria}} & \begin{scriptsize}\AR{عيشك يبارك فيك و يخليك	} \end{scriptsize}& Tunisia\\
         & \begin{scriptsize}\AR{الله يرحمه ربي معك خويا و انا لله و انا اليه راجعون	}\end{scriptsize} & Morocco\\
         \midrule
         \multirow{2}{*}{\textbf{Egypt}} & \begin{scriptsize}\AR{يلعن الكورة واليوم اللي شجعت في كورة .}\end{scriptsize}& Palestine\\
         & \begin{scriptsize}
             \AR{مرتضي صوتوا ضعيف مع كامل إحترامي مايتقارنش بنسيم مجرد مقارنة	}
         \end{scriptsize}& Tunisia\\
         \midrule
         \multirow{2}{*}{\textbf{Lebanon}} & \begin{scriptsize}\AR{حالتنا أهون من حالات كتير في الحاضر و في التاريخ . . و غيرنا كتير نجحوا .}\end{scriptsize}& Egypt\\
         & \begin{scriptsize}\AR{هههههه مين قلك أعصابي تعبانة}\end{scriptsize}& Syria\\
         \midrule
         \multirow{2}{*}{\textbf{Palestine}} & \begin{scriptsize}\AR{بما أنو آخر شهر يا ربي يكونو عاملين خصم عالفلافل	}\end{scriptsize}& Lebanon\\
         & \begin{scriptsize}
             \AR{المشكلة انه فيه ناس ماعندهم عقل عشان تعطيهم على قد عقلهم}
         \end{scriptsize}& Kuwait\\
         \midrule
          \multirow{2}{*}{\textbf{Saudi Arabia}} & \begin{scriptsize}\AR{والله ماعرف عنه بس جتني الصوره على الخاص وقلت اكيد تذكرونه	}\end{scriptsize}& Iraq\\
         & \begin{scriptsize}
             \AR{اقرا تغريدتي بالكامل وتقرا تغريدة كساب العتيبي وتعال اسال عنها وراح اجيبك	}
         \end{scriptsize}& Qatar \\
         \midrule
         \multirow{2}{*}{\textbf{Sudan}} & \begin{scriptsize}\AR{ههههههههه انت رجعتي في كلامك سمحتي سمحتي	}\end{scriptsize}& Tunisia\\
         & \begin{scriptsize}
             \AR{والله يا استاذ عوض دي عربيه	}
         \end{scriptsize}& Egypt\\
         \midrule
         \multirow{2}{*}{\textbf{Syria}} & \begin{scriptsize}\AR{هلق الاستعمار فرض علينا بس الاستحمار نحنا فينا نعمله او ما نعمله	}\end{scriptsize}& Lebanon\\
         & \begin{scriptsize}
             \AR{لاابدا ناس عندهم مبدا}
         \end{scriptsize} & Iraq\\
         \bottomrule
    \end{tabular}
    \caption{Samples of QADI for which the ADI model's predictions are also valid.}
    \label{tab:valid_FPs}
\end{table*}

\end{document}